# A DRL-based Multiagent Cooperative Control Framework for CAV Networks: a Graphic Convolution Q Network

**Jiqian Dong**
Graduate Research Assistant, Center for Connected and Automated Transportation (CCAT), and Lyles School of Civil Engineering, Purdue University, West Lafayette, IN, 47907.
Email: dong282@purdue.edu
ORCID #: 0000-0002-2924-5728

**Sikai Chen***
Postdoctoral Research Fellow, Center for Connected and Automated Transportation (CCAT), and Lyles School of Civil Engineering, Purdue University, West Lafayette, IN, 47907.
Email: chen1670@purdue.edu; and
Visiting Research Fellow, Robotics Institute, School of Computer Science, Carnegie Mellon University, Pittsburgh, PA, 15213.
Email: sikaichen@cmu.edu
(Corresponding author)
ORCID #: 0000-0002-5931-5619

**Paul (Young Joun) Ha**
Graduate Research Assistant, Center for Connected and Automated Transportation (CCAT), and Lyles School of Civil Engineering, Purdue University, West Lafayette, IN, 47907.
Email: ha55@purdue.edu
ORCID #: 0000-0002-8511-8010

**Yujie Li**
Graduate Research Assistant, Center for Connected and Automated Transportation (CCAT), and Lyles School of Civil Engineering, Purdue University, West Lafayette, IN, 47907.
Email: li2804@purdue.edu
ORCID #: 0000-0002-0656-4603

**Samuel Labi**
Professor, Center for Connected and Automated Transportation (CCAT), and Lyles School of Civil Engineering, Purdue University, West Lafayette, IN, 47907.
Email: labi@purdue.edu
ORCID #: 0000-0001-9830-2071

# ABSTRACT

Connected Autonomous Vehicle (CAV) Network can be defined as a collection of CAVs operating at different locations on a multi-lane corridor, which provides a platform to facilitate the dissemination of operational information as well as control instructions. Cooperation is crucial in CAV operating systems since it can greatly enhance operation in terms of safety and mobility, and high-level cooperation between CAVs can be expected by jointly plan and control within CAV network. However, due to the highly dynamic and combinatory nature such as dynamic number of agents (CAVs) and exponentially growing joint action space in a multiagent driving task, achieving cooperative control is NP hard and cannot be governed by any simple rule-based methods. In addition, existing literature contains abundant information on autonomous driving's sensing technology and control logic but relatively little guidance on how to fuse the information acquired from collaborative sensing and build decision processor on top of fused information. In this paper, a novel Deep Reinforcement Learning (DRL) based approach combining Graphic Convolution Neural Network (GCN) and Deep Q Network (DQN), namely Graphic Convolution Q network (GCQ) is proposed as the information fusion module and decision processor. The proposed model can aggregate the information acquired from collaborative sensing and output safe and cooperative lane changing decisions for multiple CAVs so that individual intention can be satisfied even under a highly dynamic and partially observed mixed traffic. The proposed algorithm can be deployed on centralized control infrastructures such as road-side units (RSU) or cloud platforms to improve the CAV operation.

**Keywords:** Connected Autonomous Vehicle Network, Multiagent Deep Reinforcement Learning, Information Fusion, Graphic Convolution

## INTRODUCTION

Vehicular automation is a highly anticipated technology expected to completely disrupt the transportation system (AASHTO, 2018; FHWA, 2018, 2015). The anticipated effects see improvements in mobility, efficiency, and safety across various users and communities (FHWA, 2019; Chen, 2019; Li et al., 2020; Sinha and Labi, 2007; World Bank, 2005; Du et al., 2020; Li et al., 2020a). Connectivity feature is often discussed within the context of "Internet of Things" (IoT), which enables information sharing between agents in a system (Ha et al., 2020a). The connectivity technology in autonomous driving will facilitate a CAV network, which is a collection of CAVs operating within a specific range (connectivity range) that can share observed information as well as control commands. It has been postulated that connectivity technology can further enhance the performance of the autonomous driving in 2 key aspects: cooperative sensing and cooperative maneuvering (Hobert et al., 2016; Li et al., 2020b). While cooperative sensing increases the sensing range and promotes a greater awareness of the driving environment, cooperative maneuvering refers to CAVs driving collaboratively and their motions are planned by centralized or decentralized decision processor.

Even with great potentials, various hinders have prevented the CAV networks from massive deployment. These roadblocks are not only the legislative issues but also the reliability and robustness of CAV technologies. Specifically, the technologies in design connectivity protocol and cooperative controllers are still in their infancy. Driving tasks involve complex interaction between vehicles and highly convolved decision process that cannot be easily described by the hard-coded rules (Palanisamy, 2019), and vehicle automation research has shown that there are two primary approaches to controlling a CAV system: optimization-based control and learning-based control. Optimization-based control seeks to formulate the driving tasks into a minimization (or maximization) objective function with multiple constraints and solve for control inputs. Many latest examples have successfully solved problems including CAV's trajectory planning (Yu et al., 2019), multi-platoon cooperative control (Li et al., 2019; Du et al., 2020a), joint control of CAV and traffic signals (Feng et al., 2018). However, when dealing with cooperative control among a network of vehicles, in practical, there exist vehicles entering and exiting, the number of observed vehicles as well as controlled vehicles within the network are changing accordingly. This problem is referred as dynamic-number-agent problem (DNAP). DNAP will cause failure of optimization-based control methods since they are not able to handle variable size in both inputs and outputs. Also, as the scenario complexity increases, the optimization problem tends to become over-complicated and highly non-convex that cannot be solved within linear time. Therefore, the optimization-based controller may not be able to suit into the prevail multiagent driving task.

On the other hand, by leveraging the universal functional approximation ability of Deep Neural Network (DNN), intelligent controller integrating Deep Learning (DL), Reinforcement Learning (RL) and other AI techniques have experienced great success in massive autonomous driving control tasks including: lane keeping and obstacle avoidance (S. Chen et al., 2020; El Sallab et al., 2017), lane changing decisions (Dong et al., 2020; Huegle et al., 2020), merging maneuvers (Saxena et al., 2019) roundabout driving policies (J. Chen et al., 2019), etc. The advantages of Deep Reinforcement Learning (DRL) based controllers are obvious: when the model is well trained, the inference time is fixed and fast; during the training, massive experiences can be generated from the simulated environment, making the trained model more robust and adaptive; furthermore, if well trained under properly designed settings (including states, actions, and reward function), a CAV driving algorithm can outperform human drivers because it is capable of making instantaneous and reliable driving decisions (J. Chen et al., 2019). However, DRL based controllers in most of existing literature only deal with single or fixed number of agents with both fixed-size observation and action spaces. In this paper, a novel DL based fusion method is proposed to solve the dynamic input size problem by aggregate the information from multiple sources with graphic representation. A centralized multi-agent controller is then built upon the fused information to make collaborative lane changing decisions for dynamic number of CAVs within a CAV network.

In general, DRL methods are used to model Markov Decision Process (MDP) (**Figure 1(a)**), which allows a single agent to explore the environment by observing states, taking actions and receiving rewards. However, since there exist multiple vehicles with the CAV network, the CAVs are not only interacting with the environment but also with each other, the entire system becomes a Markov Game (MG) (**Figure 1(b)**). Specifically, when all the agents are cooperative, they share a common reward function $r^1 = r^2 = \cdots = r^n = R$, then MGs degenerate into a multi-agent MDPs (MMDPs) (Boutilier, 1996). Multi-agent reinforcement learning (MARL) is applied as a generalization as RL in multiagent environment. While MARL is similar to RL, it has additional difficulties caused by interactions between agents and combinatorial action space, which render the training of such cooperative agents NP hard (Bernstein et al., 2002; Hernandez-Leal et al., 2019; Zhang et al., 2019; Ha et al., 2020b). In addressing the MARL problem, 2 kinds of control logic can be applied: centralized and decentralized. Within a centralized controller, each agent's actions are jointly computed by a common unit at each time step; in decentralized logic, each agent makes a decision for itself and treats other agents as parts of the environment. Studies show that a centralized controller that learns the effect of joint actions generally perform better compared to decentralized controllers (Yang et al., 2018).

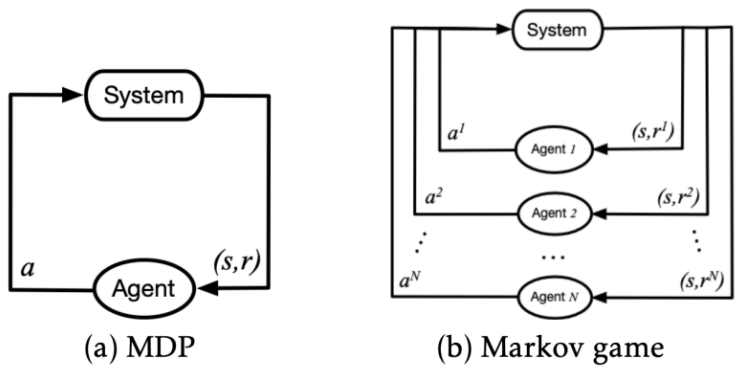

**Figure 1 Reinforcement learning settings (figures from** (Zhang et al., 2019)**)**

**CAV networks as a graph**
In a driving task, to generate safe and visionary decisions, a CAV needs not only the information of vehicles in its proximity (local information) but also the information from downstream or upstream locations (global information). In general, the local information is acquired through the onboard sensors while the global information can be obtained as a product of cooperative sensing from connectivity features. Local information is useful to constrain the CAV's short term decision such as the feasibility of making lane changes while global information enables the CAV to make future planning such as making proactive lane changing before running into a traffic jam. Therefore, comprehensively considering both local and global information is critical to enable a CAV to safely and effectively reconstruct its driving environment and generate driving decisions. Also, from the perspective of information dissemination, the HDV information are passed to CAVs from the sensor while CAVs within a certain range (namely connectivity range) can share information between each other.

      This decision dependency as well as information flow path can be modeled using a graph. Graph is a data structure with great expressive power to model a set of objects (nodes) together with their relationships (edges). Graphical representation has broad applications including in social science (Barnes, 1969), chemistry (Balaban, 1985), and transportation (Derrible and Kennedy, 2011). As shown in **Figure 2**, each node in the graph represents a vehicle and the edges represent the connection between vehicles, i.e. CAVs can obtain both information of the HDVs around them (via sensors) and the information from other CAVs (via connectivity). Therefore, the edges in the graph represent the information dissemination paths.

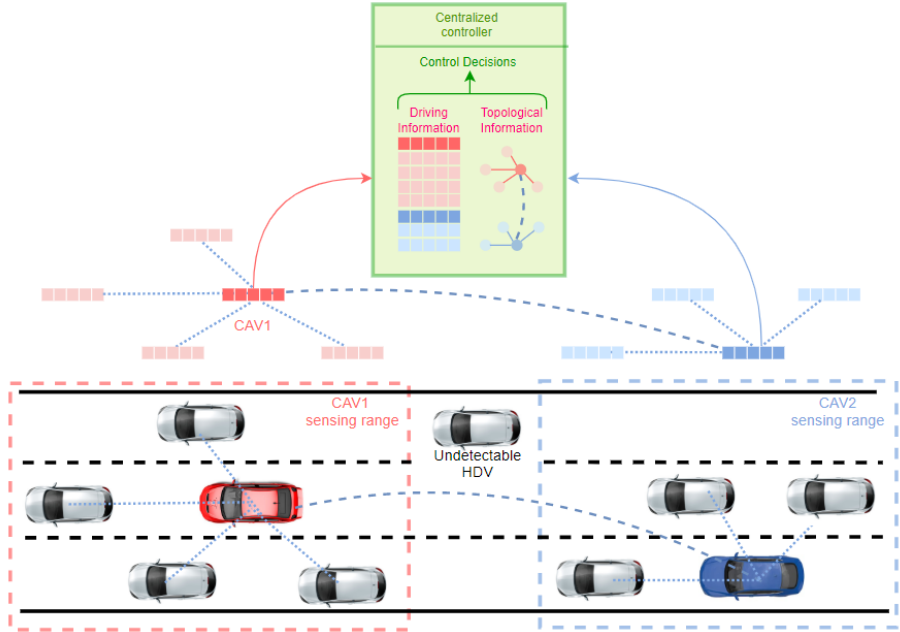

**Figure 2 Graphic representation of CAV network**

Graph Neural Network (GNN) has gained increasing popularity in various domains, including social network analysis (Qiu et al., 2018), knowledge graph (Kipf and Welling, 2019), recommender system recommendation (Fan et al., 2019), life science (Fout et al., 2017), etc. GNN has the capability to extract the relational data representations and generate useful node embeddings not only on the node features but also on the features from neighboring nodes. As the generalization of Convolutional Neural Network (CNN) to graph structure, Graphic Convolutional Network (GCN) presented by Kipf & Welling (2016) (Kipf and Welling, 2019) has great potential to aggregate the information for a clique of nodes when generating node embeddings. Both local information and global information are essential for CAV to make decisions. Thus, an explicit fusion method is needed to combine the information from theses 2 different sources. GCN, therefore is an ideal candidate for this fusion task due to its' information aggregation ability. The intuition of applying GCN for data fusion in CAVs control is that to generate the decision for CAV node, the information from its neighboring nodes including HDVs surrounding (providing local information) and other CAVs (providing global information) are incorporated contemporarily. Additionally, the weights of the GCN layers perform as an "attention" mechanism, which facilitate CAV to learn to automatically focus on the vehicles' information that are more relevant to the decision making rather than the ones that are less important based on their location relationship. For example, when a CAV agent is making a lane changing decision, proximal vehicles may have greater influence than distant vehicles, and downstream vehicles will likely have a greater impact than the vehicles upstream.

The inputs of the GCN block can be the feature matrix containing raw information (speed, location, intention, etc.) of each vehicle, and the adjacency matrix depicts the information flow topology as well as the decision dependency. The output of GCN is a node-level feature embedding map, which contains the information of both locality and global environment by fusing the raw data from 2 sources. These node embeddings can be used as necessary knowledge to draw informed and collaborative driving decisions for all the CAVs.

A good example of combining GNN and DRL is the Graph convolutional reinforcement learning (DGN) (Jiang et al., 2020). The model uses GNN as the encoder to learn abstract relational representations between agents, then feeds the representations into a policy network for actions. By jointly training the encoder and policy network, the DGN agents are able to develop cooperative and sophisticated strategies.

From Jiang et. al's ablation study, graph convolution greatly enhances the cooperation of agents (Jiang et al., 2020); such cooperation is needed in the autonomous driving task.

Inspired by DGN, this study further modifies the network so that it can produce dynamic length outputs to adapt to the autonomous driving nature. Due to the dynamic number of agents, using separate decision processors (i.e. separate Q networks) for each agent will be difficult to train together, and collaboration cannot be guaranteed (Zhang et al., 2019). Also, the number of parameters for separate Q networks will grow exponentially with the number of agents, and therefore is not scalable to be trained jointly. Further, all the CAVs should be considered homogeneous in the environment, and an explicit cooperative manner is expected for the agents. Therefore, a more efficient way to achieve desired outcomes is to use parameter sharing (Gupta et al., 2017), i.e. a shared centralized Q network to output actions for all agents. In this setting, the CAVs can be treated as probe sensors to collect data on both local and global driving environments for the centralized controller. When the number of CAVs increase, the overall traffic condition of a road segment is better understood, which will further enhance the CAV's decisions in terms of safety and systemwide mobility.

**Main contribution**
This study solves the problem associated with the highly dynamic quantity of agents, using GCN and a modified centralized deep Q network (DQN). The study contributions:
- Describe a mixed-traffic driving environment and information flow topology using a graphic representation.
- Develop a GCN based information fusion block to enforce communication and cooperation between vehicles
- Combine GCN and DQN into an end-to-end multi-agent decision processor to control the CAVs lane changing decisions for a road segment.

The remainder of this paper is organized as following: Methodology introduces detail logic behind proposed method as well as the DRL model architecture; Experiment settings describes the designed scenarios, baseline models, DRL basics including state space, action space, reward function, simulation specifics and training parameters; Results report the comparative statistics and visualization of the proposed model against multiple baselines; Conclusions summarize the study and point out the potential future works.

**METHODS**
The CAV network can be modeled as a graph whose nodes represent different vehicles (**Figure 3**). To represent mixed traffic, the graphical structure of CAV network can be further decomposed into 2 layers: local and global network based on the vehicles' spatial location and their types. The local network is a "star" graph including the "ego CAV" (the CAV in question) and its surrounding human-driven vehicles (HDVs) while the global network is constructed from all the CAVs on the road. From the perspective of information transfer, the CAV acquires both local information from the nearby HDVs through onboard sensors and acquires global information from other CAVs via connectivity channels. The graph structure can be used to define the information topology for the CAV network. Within the local "star" network, information passes from HDVs to CAVs because CAVs have sensing capabilities. From the global network, all the CAVs can share the knowledge including locally sensed information and their own information.

Thus, the study settings is a standard paradigm for multiagent planning: centralized learning but with decentralized execution (Kraemer and Banerjee, 2016). In this setting, each agent makes a decision at each timestep and the target is to achieve a same given goal for all the agents. The communication and information dissemination are modeled with GNN and the decision processor used is a Deep Q learning agent.

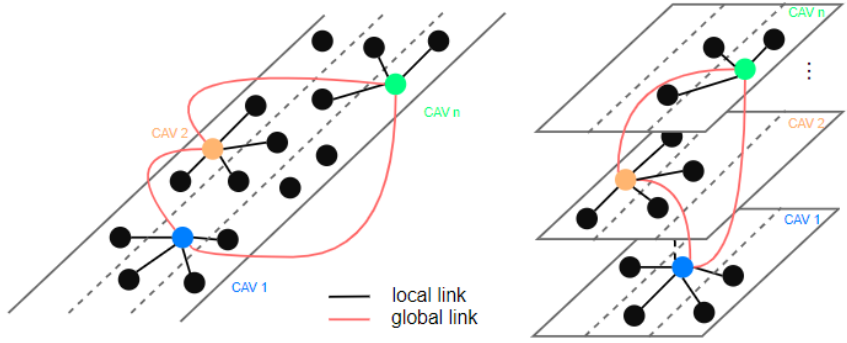

Figure 3 Graphic structure for CAV network

**Model architecture**

At each timestep $t$, $N$ vehicles including all the CAVs and the HDVs in the proximity of CAVs, are detectable. With regard to the input space of the model, at time step $t$, the state, $s_t$, is considered as a tuple of 3 blocks of information: nodes feature $X_t$, adjacency matrix $A_t$, and a mask $M_t$ documenting the index of autonomous vehicles, that is $s_t = (X_t, A_t, M_t)$. With regard to the node feature for any node $i$ (raw information for vehicle $i$ in the network), the following 4 categories are considered: speed $v_i$, location $p_i$, lane position $l_i$, and intention $I_i$. At each time step, the CAV in the neighborhood of vehicle $i$ is able to gauge the raw information of vehicle $i$ via its onboard sensors, and construct a quadruplet $x_i = (v_i, p_i, l_i, I_i)$ to represent the vehicle $i$. Since CAVs can directly share their driving information, there is no need to sense other CAVs in the vicinity of the ego CAV. All the CAVs can send surrounding vehicles' information as well as their own information back to the central control unit which concatenates all the raw information into overall node features $X_t = [x_i]_{i=1}^N$. To preserve the graph structure, during the information aggregation process, an adjacency matrix is constructed indicating the relationship between vehicles. Here, each CAV is connected with its nearby HDVs, and all the CAVs are connected. The information for both CAVs and HDVs are concatenated, and thetefore the indices for filtering out the HDV's node embeddings need to be saved, because only node embeddings for CAVs should be fed into the decision processor.

At each time step $t$, node feature matrix $X_t$ is first fed into a Fully Connected Network (FCN) encoder $\varphi$ to generate node embeddings $H_t$ in $d$ dimensional embedding space $\mathcal{H} \subset \mathbb{R}^{N \times d}$ (**Equation 1**).

$$H_t = \varphi(X_t) \in \mathcal{H} \qquad (1)$$

Then the graphic convolution is performed in the embedding space $\mathcal{H}$ for each vehicle. For each node, the GCN layer computes the node embeddings based on its own node embeddings from the encoder as well as the node embeddings for its neighboring node. In general, GCN layer computes the nodes embeddings in parallel, for all the nodes in the network, as follows:

$$Z_t = g(H_t, A_t) = \sigma(\hat{D}_t^{-1/2} \hat{A}_t \hat{D}_t^{-1/2} H_t W + b) \qquad (2)$$

Where: $\hat{A}_t = A_t + I_N$ is the adjacency matrix with self-loops for each node; $\hat{D}_t$ is the degree matrix computed from $\hat{A}$; and $\sigma$ is the nonlinear activation function such as ReLU. While there can be multiple GCN layers, the total number of layers should be restricted due to the "over-smoothing" problem (D. Chen et al., 2019). After the GCN block, the node embeddings map $Z_t$ (including both CAVs and HDVs) is obtained. Then the node embeddings for CAVs are selected out because only CAVs are controlled. Filtering can be achieved using a simple dot product of mask $M_t$ and $Z_t$:

$$Z_t^{CAV} = M_t \cdot Z_t \qquad (3)$$

The CAVs node embeddings are finally fed into a Q network $\rho$ to obtain Q values, which indicate the "goodness" of a certain action. All the neural network blocks including FCN, GCN and Q network can be summarized as $\hat{Q}$ network parameterized by $\theta$ where $\theta$ is the aggregation of all the weights.

$$\hat{Q}_\theta(s_t, a_t) = \rho(Z_t^{CAV}, a_t) \quad (4)$$

To train the model, the classic Q Learning with Experience Replay and Target Network as proposed in (Van Hasselt et al., 2016; Volodymyr Mnih et al., 2016) is applied. In order to stabilize the training, the overall neural network is trained on mini-batches randomly sampled from a replay buffer R containing transitions of $(s_t, a_t, r_t, s_{t+1})$. For each mini-batch, the objective of the training is to minimize the loss function (**Equation 5**).

$$L_\theta = \frac{1}{b}\sum_t y_t - \hat{Q}_\theta(s_t, a_t) \quad (5)$$

Where: $b$ is the batch size and $y_t = r_t + \gamma \max_a \hat{Q}_\theta(s_{t+1}, a)$. **Figure 4** presents the model layout. For each component of network, the following architecture is utilized:
- FCN Encoder $\varphi$: $Dense(32) + Dense(32)$
- GCN layer $g$: $GraphConv(32)$
- Q network $\rho$: $Dense(32) + Dense(32) + Dense(16)$
- Output layer: $Dense(3)$

Additionally, a "warming up" phase with T steps is established prior to the training in order to let the agent take random actions and fully explore the environment. This setting facilitates the agent's acquisition of adequate experiences in both successful lane changing and unsuccessful lane changing (collision), which further helps guarantee the safe lane-changing decisions. From step T+1, the training is performed by maximizing the reward and minimizing the losses as mentioned above. Algorithm 1 resents the detailed steps.

In order to train the model by batch, the input state tensors (including node features, adjacency matrices, CAV masks) and the output tensor (Q value tensor) must have the same consistent shape across each time step. To achieve this, the maximum number of vehicles in the scenario $N_{max}$ must be specified, and zero-pad both input and output tensors need to contain exactly $N_{max}$ vehicles. For example, the node features tensor for one batch has the shape $(b, N_{max}, F)$, where $F$ is the size of feature. Similarly, the adjacency matrix has the shape $(b, N_{max}, N_{max})$, and the CAV mask has the shape $(b, N_{max})$. Finally, the output Q value tensor also has the shape $(b, N_{max}, A)$, where $A$ is the size of the action space for each vehicle. During the training, the CAV mask is used to control the gradient flow by filtering out the corresponding terms for HDVs.

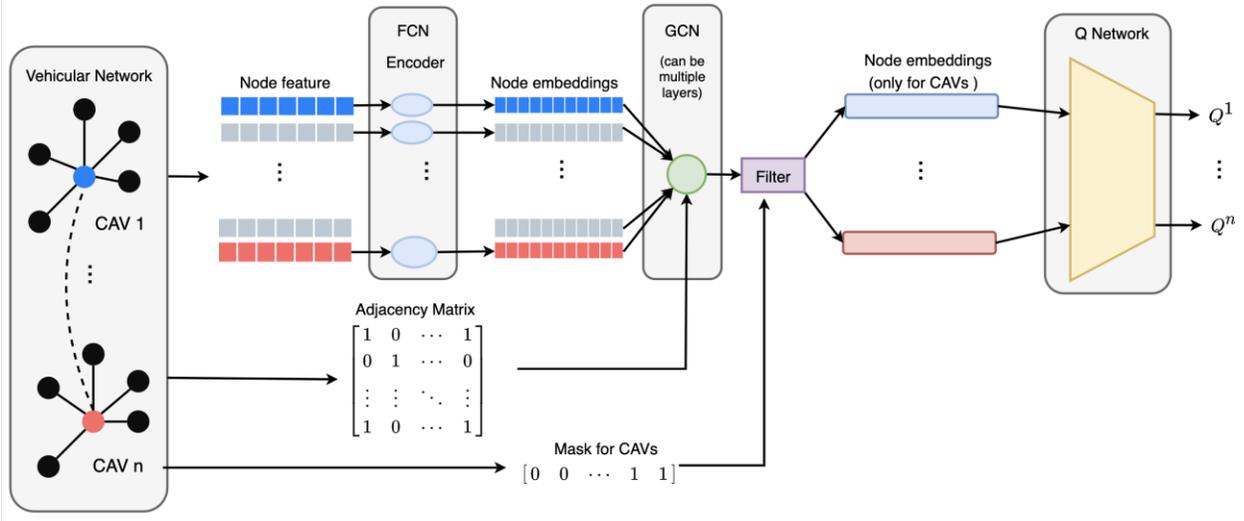

**Figure 4 Model architecture**

| Algorithm 1 | Graphic Q Learning with Experience Replay and Target Network |
|---|---|

Initialize the reply memory $R$ to capacity $N$

Initialize the weights for both Encoding block $\varphi$, graphic convolutional block $g$, Q network $\rho$ which jointly denoted as Network $\hat{Q}_\theta$ and Target Network $\hat{Q}_t = \hat{Q}_\theta$

## Warming up steps

For time step $t = 1$ to $T_1$ (warming up steps) **do**

    Take random action combination for each agent $i$: $a_t = [a_r^i]_{i=1}^n$

    Gather the transition $(s_t, a_t, r_t, s_{t+1})$

    Store the transition $(s_t, a_t, r_t, s_{t+1})$ into the memory buffer $R$

## Main training loop

For time step $t = T_1 + 1$ to $T$ (training steps) **do**

    ## Generate new samples and update memory R

    With probability $\epsilon$ select a random policy $a_t = [a_r^i]_{i=1}^n$

    Otherwise do:

    $X_t, A_t, M_t = s_t$

    Encode the raw node feature into a high dimensional feature map $H_t = \varphi(X_t)$

    Perform graphic convolution $Z_t = g(H_t, A_t)$

    Filter out the node feature for HDVs $Z_t^{CAV} = M_t \cdot Z_t$

    Compute Q values for each action combination $a_t$ $\hat{Q}_\theta(s_t, a_t) = \rho(Z_t^{CAV}, a_t)$

    Select the $a_t^* = \underset{a_t}{\mathrm{argmax}}\, \hat{Q}_\theta(s_t, a_t)$

    Execute $a_t^*$ and observe reward $r_t$ and next state $s_{t+1}$

    Store transition $(s_t, a_t^*, r_t, s_{t+1})$ into the memory buffer $R$

Set $s_t = s_{t+1}$
## Training the model at each training step
Sample random mini-batch with size b from R
For each training examples with the batch, set the target of Q value
$$y_t = \begin{cases} r_t + \gamma \max_a \hat{Q}_\theta(s_{t+1}, a_t) & \text{if } s_{t+1} \text{ is not done} \\ r_t & \text{if } s_{t+1} \text{ done} \end{cases}$$
Perform a gradient step optimizing loss function in $L_\theta = \frac{1}{b}\sum_t y_t - \hat{Q}_t(s_t, a_t)$
## Updating the Target Network
If mod(t, target updating frequency) == 0
  Set $\hat{Q}_t = \hat{Q}_\theta$

## EXPERIMENT SETTINGS
For implementing the developed framework in a simulation environment, we use an open-source simulator tool SUMO (Krajzewicz et al., 2012). In SUMO, the vehicle parameters, training environments, and vehicle control are defined and simulated. To do this, spaces (state and action) and reward functions are defined for the DRL which models the Markov Decision Process.

## Problem settings
The proposed framework is deployed within multiagent environment (**Figure 5**), a 3-lane freeway containing 2 off-ramps. All the vehicles (HDVs and CAVs) enter into the road segment from the left of the page. The color of vehicles indicates their type (CAV vs. HDV) and their intention. The CAVs are represented by colored vehicles, and their destinations are the two exit ramps. More specifically, the red CAVs endeavor to merge out from the first ramp, denoted as "merge_1" CAVs, while the green CAVs endeavor to merge out from the second ramp, denoted as "merge_2" CAVs. A major part of the reward function is based on whether or not the intention is satisfied for each vehicle, i.e., each CAV exits from their desired ramp.

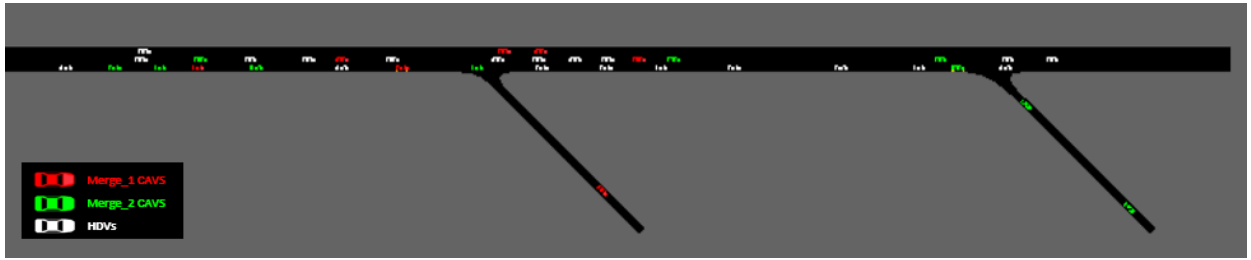

**Figure 5 Scenario settings**

## State space
At time step $t$, there are $N$ detectable vehicles and the state space contains 3 blocks: nodes feature $X_t$, adjacency matrix $A_t$, and a mask $M_t$. Each of them is computed as follows:
- Nodes features

Node features contain speed, position, location and intention, denoted as $x_i = (v_i, p_i, l_i, I_i)$, to control a relative similar scale and concatenate numerical variables with categorical variables, each of the element is manipulated as follows:
1. $v_i = \frac{v_{i-actual}}{v_{max}}$ is the relative speed for each vehicle where $v_{i-actual}$ is the actual speed of vehicle $i$ and $v_{max}$ is the speed limit for the highway segment;

2. $p_i$ describe the location, here we mainly consider the longitudinal location of the highway segment, which is normalized by the maximum length of highway, to fall between 0-1, that is: $p_i = \frac{p_{i-actual}}{l_{highway}}$
3. $l_i$ is the lane position for vehicle $i$, which is a categorical variable. "One Hot" encoding is utilized to describe the lane position, i.e. vehicles in the rightmost lane are denoted $[1, 0, 0]$, middle lane $[0, 1, 0]$ and leftmost lane $[0,0,1]$.
4. Intention $I_i$ is also a categorical variable needs "One Hot" encoding. As there are 3 intentions in total: merge out from first ramp, merge out from second ramp and go straight along the highway, $[1, 0, 0]$, $[0, 1, 0]$ and $[0,0,1]$ are set to encode 3 intentions, respectively. It may be noted that because HDV intentions are generally unobservable, a dummy value $[0,0,0]$ is used as a placeholder to keep the dimensions for such intentions.

The overall node feature map for time step $t$ is the vertical stack of all the $x_i$s. $X_t = [x_i]_{i=1}^{N}$

- Adjacency

The adjacency matrix $A_t$ is a binary $N \times N$ matrix indicating the information topology (graph structure) in CAV network. Here, an undirected graph is considered, where $A_{ij} = 1$ means vehicle $i$ and vehicle $j$ are connected. At the beginning, $A_t$ is instantiated as a zero matrix and is wired in 3 steps. First, HDVs in CAV proximity are connected to the ego CAV to represent the local link. Second, all the CAVs are connected with each other representing the global link. Third, the HDVs within an ego CAVs vicinity are connected. Since they can be sensed by the same CAV, their spatial distances are small, thus may further cause a dependency in drawing driving decisions between HDVs which may propagate to the CAV. By wiring these close HDVs together, this decisive dependency can be implicitly considered in the fusion block.

- CAV mask

CAV mask $M_t$ is used to filter out the embeddings of HDVs after the GCN fusion block. $M_t$ is constructed when assembling $x_i$s into $X_t$. It is a binary vector of length N with 1s at the indices of CAVs.

**Action space**

For each time step, every CAV has a discrete action space representing the potential actions to be undertaken, as follows:

$$a_i = \{change\ to\ left, keep\ lane, change\ to\ right\}$$

The overall action space for RL is a combinatory space aggregating all the possible combinations of individual CAV: $\mathcal{A} = \{a_i\}_{i=1}^{n}$, where $n$ is the number of CAVs at current timestep. Note, the simulator restricts agents from exiting the simulated corridor, but does not prevent vehicle collisions during the lane change process.

**Reward function**

The reward function has: 2 reward types and 2 penalty types. These are the intention reward, speed reward, lane-changing penalty, and collision penalty (**Table 1**). Intention reward is to guarantee all the CAVs can merge out from the prescribed ramp. Speed reward encourage the actions to increase the system efficiency. Specifically, in order to balance with other sources of reward and encourage cooperation, we define a "soft and smooth" intention reward at each timestep instead of giving an instant reward when the vehicle successfully merges out. As mentioned before, we denote the CAVs merging from ramp 1 as "merge_0" CAVs and the ones merging from ramp 2 as "merge_1". The intention reward is computed only on the first 2 segment: $seg_1$ and $seg_2$ on the highway (**Figure 6**) and the segment descriptions below. Denote: $p_{ij}/r_{ij}$ is the individual penalty/reward for each type $i$ CAV on the $j^{th}$ segment of road, $x_{ij}$ as the relative location of type $i$ CAV on the $j^{th}$ segment, $x_{ij}$ is restored to 0 when entering a new segment.

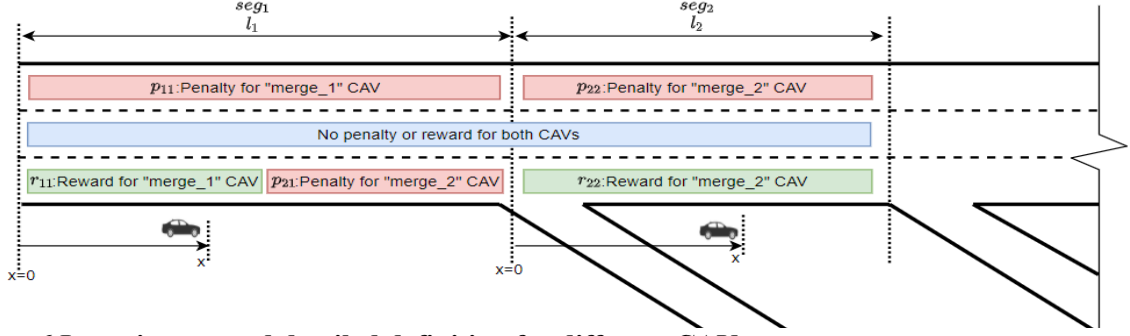

**Figure 6 Intention reward detailed definition for different CAVs**

**TABLE 1 Intention reward details**

| Segment | Category | Description | Calculation |
|---|---|---|---|
| **Seg_1** | $p_{11}$ | Penalty for "merge_1" CAV on the top lane, increases from 0-1 when approaching the ramp 1. | $p_{11} = \dfrac{x_{11}}{L_1}$ |
| | $r_{11}$ | Reward for "merge_1" CAV on the bottom lane, decreases from 1-0 when approaching ramp 1. | $r_{11} = 1 - \dfrac{x_{11}}{L_1}$ |
| | $p_{21}$ | Penalty for "merge_2" CAV on the top lane, increases from 0-1 when approaching ramp 1. | $p_{21} = \dfrac{x_{21}}{L_1}$ |
| **Seg_2** | $p_{22}$ | Penalty for "merge_2" CAV on the top lane, increases from 0-1 when approaching ramp 2. | $p_{22} = \dfrac{x_{22}}{L_2}$ |
| | $r_{22}$ | Penalty for "merge_2" CAV on the bottom lane, decreases from 1-0 when approaching ramp 2. | $r_{22} = 1 - \dfrac{x_{22}}{L_2}$ |

Then total intention reward is the submission for each CAV's individual reward or penalty on different segment, defined as:

$$R_I = \sum_{k=1}^{n} \sum_{i,j}(r_{ij} - p_{ij})\delta_{ij}^n \qquad (6)$$

Where: $\delta_{ij}^n$ is an indicator function showing whether vehicle $k$ falls into $ij^{th}$ category, $r_{ij}$ and $p_{ij}$ are the rewards and penalties for vehicle type $j$ on segment $i$, respectively. With this intention reward, a "soft" reward is set to encourage "merge_2" CAVs (**Figure 6**) to yield the bottom lane to the "merge_1" CAVs before the first ramp, and merge to the bottom lane as soon as they enter into the second segment. Also, the absolute value for reward and penalty is a function of the vehicle locations, thus encouraging the CAVs to make early decisions.

Besides intention reward, other performance metrics including efficiency, safety and comfort are considered using speed reward, collision penalty and lane changing penalty. Speed reward $R_v$ is defined as sum of the relative speed with respect to speed limit:

$$R_v = \frac{1}{n}\sum_{i=1}^{n}\frac{v_{CAV}^i}{v_{max}} \qquad (7)$$

where $v_{CAV}^i$ represent speed of $i^{th}$ CAV and $v_{max}$ is the speed limit of CAV on the road. Collision penalty $P_c$ is a fixed large positive value discourages maneuvers which may lead to a crash. Lane changing penalty $P_{LC}$ is set as a constant penalty attach to each lane changing decision, which is used to encourage CAVs to keep lane instead of frequently changing lanes.

The overall reward function is defined as:

$$R_{total} = w_1 R_I + w_2 R_V - w_3 P_c - w_4 P_{LC} \qquad (8)$$

Where: $w_1 \ldots w_4$ are weights that can be tuned by the model while duly recognizing the trade-off between the "intention satisfaction", "travel speed", "safety" and "ride comfort". When $w_1$ and $w_2$ are tuned at relatively higher levels compared to $w_3$ and $w_4$, the intention reward ($R_I$) and speed reward ($R_v$) term will dominate. This setting will encourage the vehicle to make lane changes to satisfy the intention as well as drive fast to reach destination even if the decision is error-prone and aggressive. On the other hand, an increase $w_3$ and $w_4$ over $w_1$ and $w_2$ means higher penalties for unsafe and frequent lane-change behavior. In this case, the model learns to adopt very conservative and safety-conscious behavior even at the risk of being unable to fulfil the merging intention.

**Baseline models**
In this research, 2 baseline models are implemented for comparative analysis: rule-based model and LSTM-Q network. LSTM-Q network has the similar structure as GCQ network except the fusion module. That is, instead of using graphic convolutional layers to aggregate the information for CAVs, LSTM-Q network treat information of all the vehicles as a sequence. The reason for choosing LSTM is its capability of modeling "contextual" latent information in a sequence. In the driving task, information fusion step is to get a context in the driving environment, that is, for each CAV's decision, it not only relies on CAV's own state information, but also needs the "contextual" information consists of other vehicles around it. Therefore, a LSTM layer is an ideal candidate to perform as the context extractor, which can output a sequence of "hidden states" as the node embeddings. After LSTM layer, a Q network with the same structure as GCQ model is built on top to generate actions. With regard to the number of trainable parameters, GCQ-model has 6,732 parameters while LSTM-Q has 11,331 parameters in total. As for the rule-based model, a "strategic" lane changing model which is calibrated from the actual human driving behaviors is used to control both CAVs and HDVs.

**Simulator parameters**
The driving simulation environment is defined by the parameters used in the simulator. Therefore, the simulation scenarios differ from each other in terms of the following attributes: overall traffic network features, the vehicle control which defines the interaction between vehicles, and other environment settings made specifically for training the AI, as discussed below.
- Scenario parameters

Regarding the road segment in the simulation, we adopt a 500m long, 3-lane freeway with 2 ramps at 200m and 400m locations. The speed limit for the road segment is set as 14 m/s (50km/h) for all the CAVs and 10m/s (36 km/h) for HDVs. Both CAVs and HDVs are emitted into the scenario from the left side of the highway image (Figure 6) on a random lane position and with a random initial speed. For both merge_1 and merge_2 CAVs, a constant inflow is set to be 0.1 veh/sec, while HDV inflow varies across the test scenarios.

- Vehicle control parameters

There are 2 types of vehicle control at each timestep: longitudinal and latitudinal control. Longitudinal control typically addresses the vehicle accelerations. In this work, both CAVs and HDVs are longitudinally controlled using SUMO's built-in car-following model, specifically, Trieber and Kesting's Intelligent Driver Model (IDM) (Treiber and Kesting, 2013). With regard to latitudinal control, CAVs use the output from proposed GCQ/LSTM-Q model while HDVs use the "strategic" lane changing model LC2013 (Erdmann, 2015). LC 2013 is also applied to CAVs control in the rule-based baseline model.

**Training parameters**
The model was trained using an experience reply with the first $2 \times 10^5$ transitions perform as warm-up stage before training. After the training started, transition batches with $batchsize = 32$ were sampled randomly and fed to the model. In this paper, the overall training horizon, including warm-up and actual training, is set as $8 \times 10^5$ steps (i.e., approximately 800 epochs) in total. Also, to add exploration opportunity, a simple epsilon greedy policy with a probability of 0.3 for random action is set during the training process. For the optimization parameters, *Adam* (Kingma and Ba, 2015) (a method for stochastic optimization) with an initial learning rate of $\gamma = 10^{-3}$ and a soft target model update rate $\tau = 10^{-2}$ for double Q learning are used.

**RESULTS**

The training curve on both loss and episode reward are plotted in **Figure 7**. In the training process, the first $2 \times 10^5$ steps (150 epochs) are "warming up" phase that CAVs are taking random actions for exploration. After training start, both LSTM-Q and GCQ model can converge within $8 \times 10^5$ steps. GCQ model has superior performance in terms of convergence rate and reward gained for each episode after convergence and it was observed that both models outperform the average performance of rule-based model. T was concluded that, after the training, the designed CAV controller is capable of performing lane-changing maneuvers without collision.

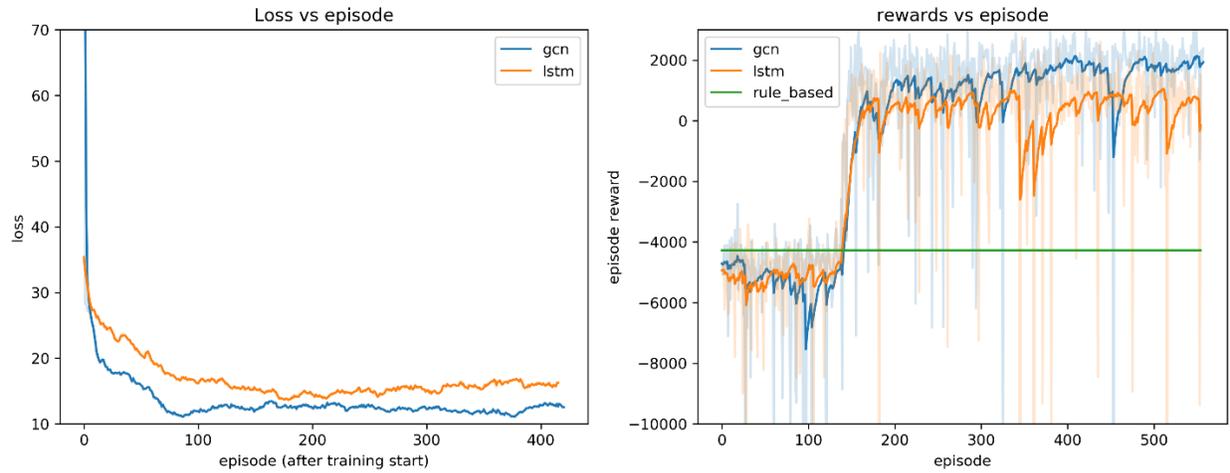

**Figure 7 Loss and rewards vs. episode**

**Comparative analysis**

The model is tested in mixed traffic different traffic densities. While training is performed under a density with a 0.2 $veh/sec$ inflow rate for HDVs, and 0.1 $veh/sec$ inflow rate for both types of CAVs (merge_1 and merge 2), the testing HDVs inflow is from 0.1-0.5 $veh/sec$. The evaluation metric we use is the mean, median and standard deviation of the episode reward acquired by running 3 models separately for 10 episodes in different traffic density scenarios. The corresponding statistics of the proposed methods vs baselines across different density scenarios are listed in **TABLE 2**, the mean and standard episode reward are plotted in **Figure 8**.

**TABLE 2 Performance comparison (episode reward) for different models in different scenarios**

| HDV Inflow Model | | 0.1 veh/sec | 0.2 veh/sec | 0.3 veh/sec | 0.4 veh/sec | 0.5 veh/sec |
|---|---|---|---|---|---|---|
| **GCQ** | mean | **3521.86** | **3537.71** | **3702.59** | **2594.94** | **4152.41** |
| | median | **3509.65** | 3635.83 | **3728.61** | **3288.57** | **4153.65** |
| | std | **491.77** | **1121.65** | 627.58 | 1815.32 | **447.03** |
| **LSTM-Q** | mean | 2075.68 | 2680.93 | 3200.40 | 2401.25 | 826.67 |
| | median | 2360.24 | **3887.89** | 3139.56 | 2951.54 | 1068.99 |
| | std | 1506.24 | 2340.38 | **479.94** | **1762.04** | 1919.53 |
| **Rule-based** | mean | -10040.15 | -6953.08 | -6762.59 | -9221.64 | -8192.19 |
| | median | -9327.27 | -6281.48 | -6227.01 | -8368.58 | -7811.41 |
| | std | 5788.20 | 4026.39 | 3678.88 | 4538.56 | 3412.66 |

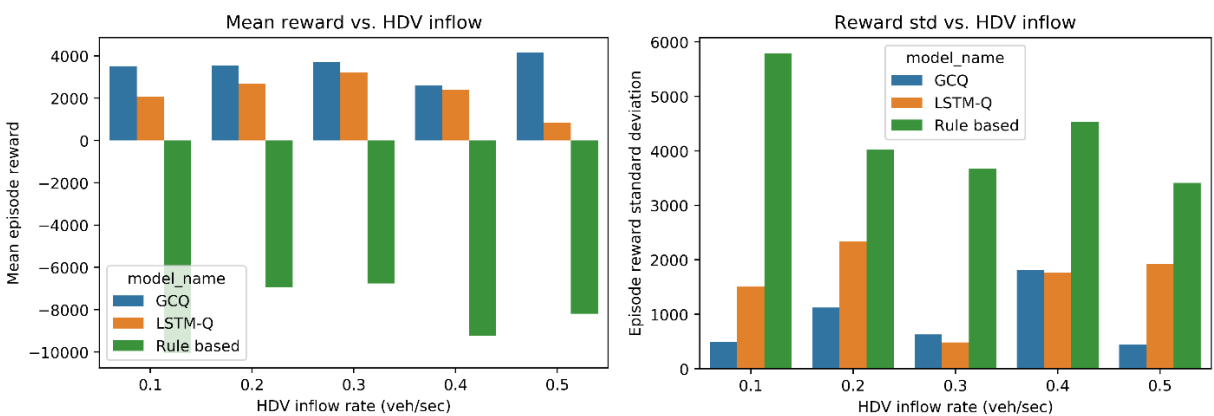

**Figure 8 Mean and standard deviation of episode rewards across different traffic density**

As shown in both **TABLE 2** and **Figure 8**, the proposed GCQ model outperforms LSTM-Q model in all the experimental scenarios of different traffic density, and both DRL-based models outperform the Rule-based model (LC-2013) by significant margins. The visualization in the simulated environment demonstrates that both GCQ model and LSTM-Q model can guide the CAVs to yield the lane for other vehicles as well as merge out from the prescribed ramp with no congestion and collision. This is a result of ideal cooperative behavior. GCQ models make more consistent decisions with a smaller variance while LSTM-Q models sometimes fails to guide the CAV to make the right decisions. That is because for LSTM models, the order of input sequence matters when generating the context information, which is not always true in driving task because the sequence of the neighboring vehicles should not influence the decision if their information is incorporated in the decision-making process. Therefore, when performing information fusion, a superior strategy, clearly, is to use a "permutation invariant" model such as the graphic convolutional neural network.

The rule-based model can guarantee all the vehicles merge out successfully, albeit with very low efficiency in certain cases. One clear problem for the rule-based model is shown in **Figure 9 (a)**: merge_2 (green) CAVs capture the bottom lane before the first ramp and block the way of the merge_1(red) CAVs. Under this situation, merge_1 CAVs must wait at the intersection until the bottom lane is clear to merge, therefore will lead to a traffic jam at the intersection. This can be avoided if merge_1 CAVs are cooperative and actively yield the bottom lane to merge_2 (red) CAVs before the first ramp. Another certain flaw is shown in **Figure 9 (b)**, when the merging CAVs (merge_2, green) fail to reach the bottom lane before the ramp and has to wait at existing position until the bottom lane is clear for merging out. The situation can be alleviated if CAVs take proactive actions to reach the exiting lane before. These defective behaviors of rule-base model will not only reduce the total efficiency of the system, but also may cause severe traffic accidents in the real world.

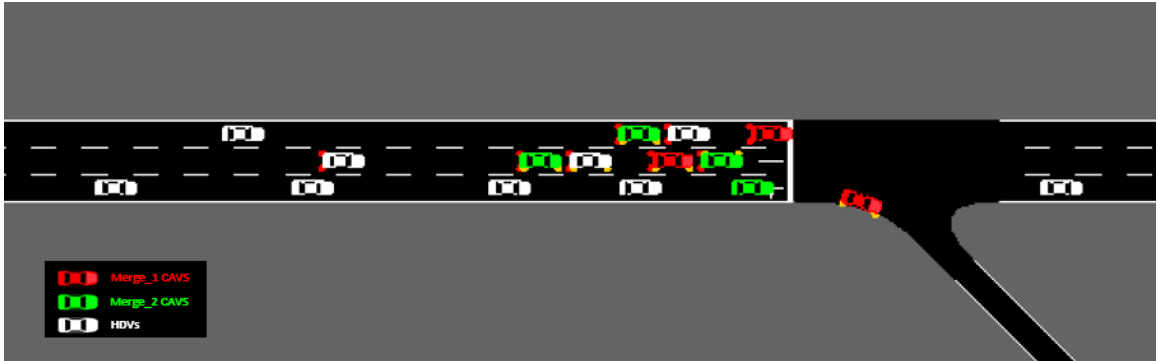

**(a) Target lane captured by undesired CAVs at Ramp 1**

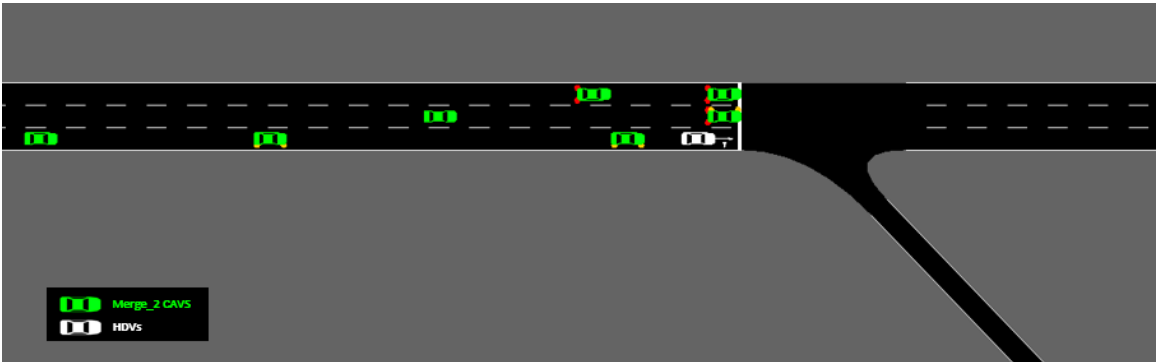

**(b) Target lane captured by HDVs at Ramp 2**

**Figure 9 A demonstration of "flaw" cases for rule-based model**

**DISCUSSION AND CONCLUDING REMARKS**

In this paper, a DRL-based model combining GCN and deep Q network (GCQ) is proposed to control multiple CAVs within a CAV network to make collaborative lane changing decisions. From the CAVs operation perspective, the proposed model not only enables the CAVs to make successful lane changes to satisfy their individual intention of merging out from the prescribed ramps but can also guarantee the safety and efficiency. As part of efforts to achieve this overarching objective, this paper also demonstrates the efficacy of the proposed model in: (a) resolving dynamic-number-agents problem (DNAP) specifically for the driving task with high model flexibility; (b) fusing information acquired by cooperative sensing on both local and global information; (c) making safe and collaborative decisions based on the fused information; (d) having enough robustness across scenarios with different traffic density and making consistent decisions without the need of retraining the model. For a comparative evaluation, the proposed GCQ model was compared with 2 baseline methods including classic "context extractor" LSTM-Q network and the traditional rule-based model calibrated from the human driving experiences. As a result, the proposed model outperforms both baselines at a significant margin. Specifically, compared to LSTM-Q model, GCQ model has much less parameters and can be trained much faster, indicating the GCN layer has the capacity to efficiently fuse essential information to generate decisions. This model can be useful when developing CAV-related centralized control unit such as roadside units (RSUs) or cloud computing platforms.

In this research, all the decisions are made instantly based only on the information at current timestep, but moving forward to the future work, with the help of connectivity and storage system, research may find it worthwhile to consider temporal information including historical data on the vehicle position, speed, and acceleration at different locations. Temporal information could serve as an indicator of looming adverse traffic conditions such as accidents, workzones, potholes etc., which may further enable the CAVs to make longer term proactive evasive decisions. By contemporarily considering "current" and "past", the CAV can output smoother and comfortable decisions series for the "future". Another potential direction for future research is to maximize the utility of all vehicles in the entire corridor rather than the CAVs utility through collaboratively controlling the CAVs. Examples such as using CAVs to mitigate traffic congestion, promote traffic string stability and reduce fuel consumption or emissions can be investigated with the proposed GCQ model.

**ACKNOWLEDGMENTS**
This work was supported by Purdue University's Center for Connected and Automated Transportation (CCAT), a part of the larger CCAT consortium, a USDOT Region 5 University Transportation Center funded by the U.S. Department of Transportation, Award #69A3551747105. The contents of this paper reflect the views of the authors, who are responsible for the facts and the accuracy of the data presented herein, and do not necessarily reflect the official views or policies of the sponsoring organization.

**AUTHOR CONTRIBUTIONS**
The authors confirm contribution to the paper as follows: all authors contributed to all sections. All authors reviewed the results and approved the final version of the manuscript.